\documentclass[journal]{IEEEtai}

\usepackage[colorlinks,urlcolor=blue,linkcolor=blue,citecolor=blue]{hyperref}

\usepackage{color,array}

\usepackage{graphicx}

\IEEEoverridecommandlockouts
\usepackage{cite}
\usepackage{amsmath,amssymb,amsfonts}
\usepackage{algpseudocode}
\usepackage{algorithm2e}
\usepackage{multirow}
\usepackage{graphicx}
\usepackage{textcomp}
\usepackage{xcolor}
\usepackage{booktabs,siunitx}
\usepackage[caption=false,font=normalsize,labelfont=sf,textfont=sf]{subfig}
\usepackage{stfloats}
\usepackage{url}
\usepackage{verbatim}
\usepackage{makecell}

\begin{document}

\title{FedFiTS: Fitness-Selected, Slotted Client Scheduling for Trustworthy Federated Learning in Healthcare AI\\
}

\author{\IEEEauthorblockN{Ferdinand Kahenga$^\dag$, Antoine Bagula$^\dag$,  Sajal K. Das$\ddag$, and Patrick Sello$^\dag$\\}
$\dag$Department of Computer Science, University of the Western Cape,
Cape Town, South Africa\\

$\ddag$Department of Computer Science,
Missouri University of Science and Technology, Rolla,
Missouri, USA \\
E-mail: 
ferdinandkahenga@esisalama.org, abagula@uwc.ac.za, sdas@mst.edu, 3919362@myuwc.ac.za
}


\maketitle

\begin{abstract}
Federated Learning (FL) has emerged as a powerful paradigm for privacy-preserving model training, yet deployments in sensitive domains such as healthcare face persistent challenges from non-IID data, client unreliability, and adversarial manipulation. This paper introduces \textit{FedFiTS}, a trust- and fairness-aware selective FL framework that advances the FedFaSt line by combining \emph{fitness-based client election} with \emph{slotted aggregation}. FedFiTS implements a three-phase participation strategy—free-for-all training, natural selection, and slotted team participation—augmented with dynamic client scoring, adaptive thresholding, and cohort-based scheduling to balance convergence efficiency with robustness. A theoretical convergence analysis establishes bounds for both convex and non-convex objectives under standard assumptions, while a communication-complexity analysis shows reductions relative to FedAvg and other baselines. Experiments on diverse datasets—medical imaging (X-ray pneumonia), vision benchmarks (MNIST, FMNIST), and tabular agricultural data (Crop Recommendation)—demonstrate that FedFiTS consistently outperforms FedAvg, FedRand, and FedPow in accuracy, time-to-target, and resilience to poisoning attacks. By integrating trust-aware aggregation with fairness-oriented client selection, FedFiTS advances scalable and secure FL, making it well suited for real-world healthcare and cross-domain deployments.
\end{abstract}

\begin{IEEEkeywords}
Federated Learning, Selective FL, Trust-Aware Aggregation, Fairness in FL, Convergence Guarantees, Security, Healthcare AI
\end{IEEEkeywords}

\vspace{-0.15in}
\vspace{-0.05in}

\section{Introduction}

The digitisation of healthcare and advances in artificial intelligence (AI) have reshaped medical data usage, enabling improved decision-making. Federated Learning (FL) represents a pivotal development, allowing collaborative model training across institutions without compromising data privacy, making it a crucial aspect in medical contexts.

However, FL in healthcare must address challenges related to trust, transparency, security, and fairness. The lack of interpretability often hinders trust in AI-assisted diagnoses, while dataset heterogeneity may lead to bias and unfairness, affecting patient outcomes. Security threats such as adversarial attacks and unreliable client participation further compromise system robustness.

To mitigate these issues, this work proposes \textit{FedFiTS} (Federated Learning using Fittest Parameters Aggregation and Slotted Clients Training), a novel trust-aware and security-enhanced FL algorithm. It incorporates a selective client participation strategy, evaluating clients based on performance, data quality, and reliability, and employs a fitness-aware aggregation mechanism to ensure that only optimal and unbiased models influence the global training process. 

\textit{FedFiTS} extends our earlier work on the FedFaSt algorithm~\cite{FedFaSt} by introducing dynamic client scoring, adaptive slot-based selection using performance fluctuation tracking, and trust-aware aggregation. These enhancements strengthen robustness against adversarial behavior and improve fairness, making the approach more suitable for real-world healthcare deployment.


Key challenges faced in healthcare-oriented federated learning include:  

\begin{itemize}
    \item \textbf{Trust and Explainability:} Healthcare professionals require AI models that are not only accurate but also \textit{interpretable and reliable}. Black-box FL models can lead to scepticism and hesitation in clinical settings.  
    \item \textbf{Bias and Fairness in Medical AI:} If federated models are trained on non-representative data, they may produce \textit{unfair or inaccurate predictions} that disproportionately impact specific patient populations.  
    \item \textbf{Security Risks in Federated Learning:} The decentralised nature of FL makes it vulnerable to \textit{data poisoning attacks, adversarial manipulations, and unreliable client contributions}, all of which affect overall model performance.  
\end{itemize}


To address these challenges, \textbf{FedFiTS} introduces:  

\begin{itemize}
    \item \textbf{Trust-Aware Model Aggregation:} Ensuring that \textit{only high-quality, reliable, and unbiased clients} contribute to the global FL model.  
    \item \textbf{Security-Enhanced Federated Learning:} Improving resilience against \textit{data poisoning, adversarial attacks, and unreliable client participation} through \textit{threshold-based client selection}.  
    \item \textbf{Fair and Inclusive AI for Healthcare:} Reducing bias in medical AI by dynamically adjusting \textit{client selection strategies to enhance fairness across diverse patient demographics}.  
\end{itemize}

\vspace{2pt}
\textit{FedFiTS} is an innovative selective federated learning algorithm that advances FL through a three-pronged strategy: i) Free-for-all (FFA) training, assessing the client model training efficacy; ii) Natural selection (NAT), electing the most proficient clients for the FL model; and iii) Slotted team participation (STP), wherein the clients and parameter aggregation undergo training over successive rounds. The FFA accommodates all clients in training, while the NAT strikes a balance between data quality and training performance to cherry-pick optimal clients. The STP assigns a designated group of clients a predetermined number of rounds for training and aggregation before FFA reassessment in contrast to the round-based systems~\cite{Fedavg},~\cite{clientSelectStatPow}. This innovation, nurturing a consistent team to spearhead training and aggregation rounds, holds the potential to expedite the FL execution time. Moreover, FedFiTS augments the accuracy and convergence probability through NAT-driven selection process. The performance of FedFiTS is compared with FedAVG and two selective FL algorithms, FedRand and FedPow, leveraging Pneumonia Chest X-rays dataset~\cite{xrays}.


The rest of this paper is structured as follows: Section II surveys related work on selective and secure FL for healthcare; Section III introduces \textit{FedFiTS} and its client-selection mechanism; Section IV presents the theoretical analysis and convergence results; Section V reports experiments on real-world medical datasets against strong baselines; Section VI concludes with future directions in trust-aware, fairness-oriented federated learning.

\section{Related Work} \label{sec:2}

The existing literature related to our proposed work is divided into two main categories: selective FL and security in FL, as described below.

\subsection{Selective FL}
In Federated Learning (FL), it is crucial to address the intricate challenges posed by the heterogeneity in both data and system configurations. While the fundamental FedAVG algorithm~\cite{Fedavg} serves as a baseline, an array of innovative strategies has emerged to combat statistical data heterogeneity. For instance, FedProx~\cite{FedProx} introduces a novel regularization term at the client side, aligning local and global model updates. Distinctly, FedNova~\cite{Fednova} normalizes and scales local updates based on varying local step counts prior to global model updates, mitigating naive aggregation. On-device approaches, exemplified by SCAFFOLD~\cite{scaffold}, emulate centralized updates by employing corrective terms on each device.
Mitigating the challenges of non-identically and independently distributed (non-IID) data, methods involving data sharing from the server to the clients, data augmentation, and generative adversarial networks (GANs)~\cite{federatedNonIidSurvey} have also surfaced. 

Complementary to these advancements, selective federated learning techniques have emerged~\cite{clientSelecRes1,clientSelecRes2,clientSelecRes3} to accelerate convergence and detect anomalous behaviors stemming from both system and statistical heterogeneity. A notable instance is found in~\cite{clientSelectStatAnomal}, where the clients exhibiting higher local losses are designated as the training team in the subsequent round. FedRand, a selective variant of FedAVG~\cite{Fedavg}, randomly selects $m = cK$ clients, where $0 < c < 1$ and $K$ is the total number of clients. In a different vein, a sampling strategy is adopted in~\cite{clientSelectStatProb}, choosing the clients proportionally to their data fractions.

Refinements to client selection mechanisms persist. In~\cite{clientSelectWeightsBased}, the divergence between each client and a reference client at the server (client0) is leveraged such that those displaying the least divergence are chosen for further rounds. In~\cite{clientSelectStatAccBased}, a ranking mechanism grounded in server-side accuracy evaluation designates clients for participation. A distinct paradigm unfolds in~\cite{clientSelectStatAnomal}, as the clients' scores are computed based on their performance on an observed dataset. The server's dataset facilitates model evaluation, with the resulting loss contributing to the score. In our prior work~\cite{FedFaSt}, we introduced \textbf{FedFaSt}, a selective federated learning algorithm based on a three-step strategy: Free-for-All (FFA) training to evaluate client performance, Natural Selection (NAT) to select the fittest clients based on a combined score of data quality and model proximity, and Slotted Team Participation (STP) to allow selected clients to train over multiple rounds before reevaluation. FedFaSt demonstrated improved convergence speed, accuracy, and resilience to data poisoning compared to FedAVG, FedRand, and FedPow. However, the original FedFaSt employed static hyperparameters and lacked dynamic client evaluation mechanisms. To address these limitations, the present work introduces \textbf{FedFiTS}, which extends FedFaSt by incorporating adaptive performance fluctuation tracking, dynamic weighting of data quality versus model alignment, and trust-aware client selection mechanisms. These enhancements improve the robustness, fairness, and usability of federated learning systems, particularly in sensitive healthcare environments.

\subsection{Security in FL}

In terms of security, FL systems face challenges related to targeted and poisoning attacks. 
These attacks, known as data and model poisoning, occur when an adversary successfully alters either the data or the parameters of a model during local training on a client.
Data poisoning involves altering or corrupting data or labels, or injecting fabricated data into a client’s training dataset. 
The objective of such an attack is to introduce bias, compromise the training process, and affect the final model's accuracy or integrity \cite{attack1}. 
Data poisoning can take various forms, such as altering labels (label flipping), injecting harmful data (data injection), or embedding hidden triggers (backdoor attacks) to manipulate model behavior \cite{attack2}.

To mitigate these attacks, various defense approaches have been proposed. 
On the server side, received client models can be evaluated individually using an independent dataset before aggregation~\cite{clientSelectStatAccBased}. 
Other defense techniques rely on calculating gradient differences to identify clients whose updates are statistically different from the others, excluding them when necessary (e.g., Krum~\cite{krum}). 
Another approach does not entirely exclude byzantine clients but instead reduces the impact of their outlier parameters. 
This can involve removing the outliers during the aggregation phase (e.g., Trimmed Mean~\cite{Tmean}) or using the median of the parameters (e.g., Median-based Filtering~\cite{MedianFiltering}), as the median is less sensitive to outliers.

Trust mechanisms like blockchain, secure multiparty computation, differential privacy, and trusted execution environments have consistently upheld model reliability and data privacy~\cite{WCFLMBPP, BBFLMSH, AP2FL}, achieving results comparable to centralized learning while supporting compliance with healthcare regulations~\cite{FLTrust, LFWSPrivacy}. The \textit{FedFiTS} framework exemplifies this, using a phased client selection strategy—FFA, NAT, and STP—to prioritise high-performing participants, thus improving outcomes and mitigating data heterogeneity issues. The importance of adaptive client selection is echoed in studies using dynamic and resource-aware participation strategies~\cite{BBFLMSH, TADRLFL}, affirming the value of trust-aware FL in scalable, secure healthcare AI.

\begin{table}[ht]
\vspace{-2mm}
\centering
\caption{Trust mechanisms in federated learning for healthcare.}
\label{table:trust_mechanisms}
\renewcommand{\arraystretch}{1.1}
\footnotesize
\begin{tabular}{|p{2.3cm}|p{2.7cm}|p{0.8cm}|p{1.2cm}|}
\hline
\textbf{Trust Mechanism Type} & 
\textbf{Accuracy Improvement} & 
\shortstack{\textbf{Privacy}\\\textbf{Score}} & 
\shortstack{\textbf{Implement.}\\\textbf{Complexity}} \\
\hline
Blockchain & 10--20\% improvement in several metrics & High & High \\ 
Uncertainty-aware learning & Outperforms FedAvg on out-of-distribution data & Medium & Medium \\ 
Trust scores based on root dataset & Low testing error rates & Medium & Low \\ 
Adaptive differential privacy & High model accuracy & High & Medium \\ 
Cyclic knowledge distillation & 10\%+ improvement & Medium & Medium \\ 
Dynamic weight-based fusion & 13.28\% improvement & Medium & Medium \\ 
Reputation and Trust-based technique & Improved model effectiveness & Medium & Medium \\ 
Federated learning (general) & Comparable to centralized learning & High & Medium \\ 
FedAvg algorithm & 75\% detection accuracy & High & Low \\ 
Graph convolutional networks & No mention found & High & High \\ 
Particle Swarm Optimization & 97\% accuracy & High & Medium \\ 
Deep Reinforcement Learning & 97.3--99.4\% accuracy & Medium & High \\ 
Decentralized functional encryption & Better model accuracy & High & High \\ 
Trusted Execution Environments (TEE) & Outperforms existing methods & Very High & High \\ \hline
\end{tabular}
\vspace{-3.5mm}
\end{table}
Table~\ref{table:trust_mechanisms} summarizes trust-enhancement mechanisms in federated learning, including blockchain auditing, reputation-based aggregation, and fairness incentives. These approaches, while effective, often incur high overhead, static scoring, or narrow data focus. \textit{FedFiTS} complements them with a dynamic $\alpha$--$\beta$ scoring model that balances data quality, performance, and fairness. By adaptively selecting participants, it offers a lightweight and scalable alternative to heavy-weight methods, while also integrating with existing frameworks—for example, by leveraging external trust ratings or reducing blockchain verification overhead through slotted training. Thus, \textit{FedFiTS} operates both as a standalone trust mechanism and as a complement to the approaches in Table~\ref{table:trust_mechanisms}.

\subsection{What gap does \textsc{FedFiTS} fill?}
Table~\ref{tab:fedfast_vs_fedfits} highlights three persistent deployment gaps left by \textsc{FedFaSt} and how \textsc{FedFiTS} closes them:

\textit{Gap 1: Selection bias and client starvation.} Single-metric, top-$K$ selection repeatedly favours the same ``fit'' clients, sidelining minority cohorts and wasting rounds when utility drifts. \textit{\textsc{FedFiTS}:} replaces scalar fitness with a \emph{multi-objective} score (utility, gradient coherence, diversity, staleness/energy, DP) and uses explore–exploit policies plus participation floors/quotas to prevent starvation while keeping throughput high.

\textit{Gap 2: Brittle, round-synchronous scheduling.} Fixed slotted teams and rigid cadence amplify staleness, penalise intermittent clients, and stall when gradient variance changes across phases. \textit{\textsc{FedFiTS}:} introduces \emph{adaptive} team slots (size/length tuned by gradient variance and plateau tests), late-arrival handling, and optional async catch-up, reducing idle time and staleness without sacrificing stability.

\textit{Gap 3: Fragile aggregation under non-IID, outliers, and privacy noise.} Vanilla FedAvg/FedAdam can be skewed by majority domains, poisoned by outliers, and destabilised by DP noise; fairness is often implicit and unreported. \textit{\textsc{FedFiTS}:} adopts a \emph{two-stage} aggregation (slot-internal $\rightarrow$ cross-slot) with robust fallbacks (median/Krum), optional similarity-aware head mixing, trust decay and gradient-cosine outlier checks, minority-cohort reweighting, and a non-convex \emph{stationarity} bound that makes residuals (mixing, drift, label noise) explicit.

\noindent Taken together, \textsc{FedFiTS} upgrades selection, scheduling, and aggregation from heuristics to \emph{accountable, fairness-aware} mechanisms that remain stable and efficient under real non-IID participation and resource constraints.

\begin{table}[!t] \centering \caption{Key differences between \textsc{FedFaSt} and \textbf{FedFiTS}.} \label{tab:fedfast_vs_fedfits} \renewcommand{\arraystretch}{1.12} \footnotesize \begin{tabular}{p{2.1cm} p{2.5cm} p{2.6cm}} \hline \textbf{Aspect} & \textbf{\textsc{FedFaSt} (prior)} & \textbf{\textsc{FedFiTS} (this work)} \\ \hline Client selection & Heuristic scalar fitness; top-$K$ per round & \textit{Multi-objective} fitness (utility, grad. coherence, diversity, staleness/energy, DP); explore–exploit to avoid starvation \\ Team slots / scheduling & Fixed slotted teams; round-based cadence & \textit{Adaptive} team slots (size/length by gradient variance/plateau tests); late-arrival handling; optional async catch-up \\ Aggregation & FedAvg/FedAdam (+ optional trust) & Two-stage: \textit{slot-internal} $\rightarrow$ \textit{cross-slot}; robust fallback (median/Krum); optional similarity-aware head mixing; secure-agg ready \\ Fairness & Implicit & Participation floors/quotas; minority-cohort reweighting \\ Robustness & Basic quality gating & Trust decay; gradient-cosine outlier checks; drift guards \\ Convergence view & Empirical & Non-convex \textit{stationarity} bound with residuals for mixing/clustering/label-noise \\ Overhead & Low & Low–moderate (selection + robust agg), offset by fewer wasted rounds \\ When preferable & Stable/homogeneous FL & Non-IID, unreliable participation; fairness/robustness needed \\ \hline \end{tabular} \end{table}

\section{Fittest Selection, Slotted Client Scheduling for Trusthworthy FL Model} \label{sec:3}
\vspace{-0.05in}

\subsection{FedFiTS Problem Formulation}
FedFiTS is a selective federated learning algorithm that relies on the following metrics to select the fittest clients in the aggregation and training processes. 

\vspace{2pt}
\noindent{\bf Quality of Learning (QoL): $\theta_k$ \& $score_{k}$ } 

\begin{figure}[htbp]
    \centering
    \subfloat[{\fontsize{8}{12}\selectfont$M$ and $\theta$ representation}]{%
        \includegraphics[width=0.22\textwidth]{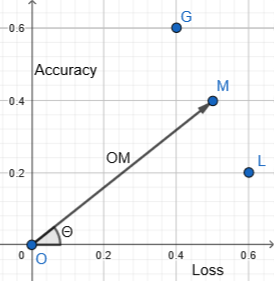}
        \label{fig:pointM}
    }
    \hfill
    \subfloat[{\fontsize{8}{12}\selectfont FedFiTS clients selection}]{%
        \includegraphics[width=0.22\textwidth]{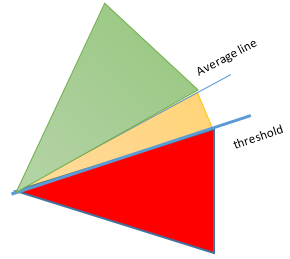}
        \label{fig:solutionRep}
    }
    \caption{Showing client performance and FedFiTS client selection}
    \label{et}
\end{figure}

Let us consider the global model $w(t-1)$ and the local model $w_k(t)$ at round $t$ on client $k$. Let $GL_k$, $GA_k$, $LL_k$, and $LA_k$ denote the training performance metrics in terms of accuracy and loss. $GL_k$ and $GA_k$ represent respectively the loss and accuracy  of the global model $w(t-1)$ while $LL_k$ and $LA_k$ are loss and accuracy of the local model $w_k(t)$ using the local dataset $D_k$. To measure the gap between models $w(t-1)$ and $w_k(t)$,  we consider  $GL_k$, $GA_k$, $LL_k$, $LA_k$ to be the coordinates of points  $G(GL_k, GA_k)$ and $L(LL_k, LA_k)$ in a two-dimensional plane where the loss is represented by the $X$-axis while the accuracy by the 
$Y$-axis. As shown in Fig.\ref{fig:pointM}, a middle point $M$ located on the OM segment is defined by 
M($\frac{GL_k+LL_k}{2},\frac{GA_k+LA_k}{2}$). Owing to the difference between accuracy $[0,1]$ and loss $]0,+\infty[$ domains and for the sake of standardization between the two metrics, we propose to use the angle $\theta_k$ as a measure of the distance between $M$ and the loss unit vector $[1,0]$.
\vspace{-0.05in}
\begin{equation} \label{eq:theta_calculation}
\theta_k = arcos(\frac{GL_k+LL_k}{\sqrt{(GL_k+GA_k)^2+(LL_k+LA_k)^2}})
\end{equation}
Eq.~(\ref{eq:theta_calculation}) denotes the gap between client $k$'s model performance and the server's model performance. When comparing the clients $k$ and $k+1$, if $\theta_k>\theta_{k+1}$, client $k$'s performance is closer to server's global model $w(t-1)$ than that of $k+1$. 

Let $n$ be the size of the global dataset (including all clients), $n_k$ the local client $k$'s dataset size and $K$ the total number of clients. The expression 
$q_k=\frac{n_k}{n}$
represents the data quality measure for client $k$ such that $\sum_{k=1}^{K}{\frac{n_k}{n}=1}$.
The client's fitness to enter the training and 
aggregation team, referred to as the {\it training performance metric}, is defined by a weighted sum of the client's performance measure expressed by Eq.~(\ref{eq:score_equation}) and data quality measure defined above.
For client $k$ at round $t$, the training performance metric is defined by:
\vspace{-0.05in}
\begin{equation} \label{eq:score_equation}
\\score_{k}(t)= \alpha q_k + (1-\alpha)\theta_k
\end{equation} 
where $\alpha$ is a hyper-parameter controlling the trade-off between the quality of data held by the client ($q_k$) and  performance ($\theta_k$).

\vspace{2pt}
\noindent{\bf Clients' Fitness Threshold: $threshold(t)$.}

FedFiTS uses a threshold-aware fitness assessment of clients. During the current round, the clients with 
performance above a threshold are elected as fit to enter the 
training and aggregation team for the next slot; and those with 
performance below the threshold are left aside for fitness 
re-evaluation at the end of the current slot. As depicted in 
Fig.~\ref{fig:pointM}, the segment OM represents a trade-off in 
terms of performance between the global ($G$) and local ($L$) models. It should therefore be used to guide the definition of a threshold to be applied to the client's ranking/scoring to differentiate between the fittest and unfit clients. Such a threshold is given by 
\vspace{-0.05in}
\begin{equation} \label{eq:threshold_equation}
\\threshold(t) = \frac{\Sigma_{k=1}^{K}{score_k}}{k} \times (1-\beta)
\end{equation} 
where the hyper-parameter $\beta$ smooths the selection 
process to enable border-line clients just below the average line in Fig.~\ref{fig:solutionRep} to be selected in the training team. 
While  $\alpha$ and $\beta$ 
control the 
clients' participation in the training and aggregation processes, the  
threshold line in Fig.~\ref{fig:solutionRep} indicates how close/far the clients' selection scheme is to the average solution where the threshold line coincides with the average line.  

Fig.~\ref{fig:solutionRep} provides a pictorial 
representation of the FedFiTS clients' selection process where i) the green clients over the average line are controlled by hyper-parameter $\alpha$, 
ii) the red clients are in the waiting zone for re-evaluation, and iii) 
the yellow clients located between the green and red clients are just below the average line and controlled by $\beta$.  

\vspace{2pt}
\noindent{\bf Training Slots: MSL, PFT, and \textit{p(t)}}

One of the critical challenges in federated learning is managing how often and which clients participate during training rounds. Frequent client switching can cause instability and degrade fairness, while infrequent switching might prevent the inclusion of diverse updates from heterogeneous clients. To strike a balance, FedFiTS introduces a \textbf{slot-based training mechanism} that dynamically adjusts the \textbf{duration} of client team participation based on their performance.

We define three key components in this mechanism:

\begin{itemize}
    \item \textbf{MSL (Maximum Slot Length)}: the maximum number of consecutive rounds a selected team of clients can participate before a mandatory reassessment.
    \item \textbf{PFT (Performance Fluctuation Threshold)}: a tolerance threshold that defines how many consecutive rounds of declining team performance are allowed before triggering a reselection.
    \item \textbf{$p(t)$}: a dynamic counter at round $t$ that tracks how many successive rounds the team’s overall performance has decreased.
\end{itemize}

To determine team stability, we observe the evolution of the overall performance metric $\theta(t)$. If performance declines compared to the previous round ($\theta(t) < \theta(t - 1)$), we increment the counter $p(t+1) = p(t) + 1$; otherwise, we reset it to zero:

\begin{equation}
p(t+1) =
\begin{cases}
p(t) + 1, & \text{if } \theta(t) < \theta(t - 1) \\
0, & \text{otherwise}
\end{cases}
\end{equation}

A reselection is triggered when either of the following conditions is met:
\begin{enumerate}
    \item The number of consecutive declining rounds exceeds the threshold: $p(t+1) \geq \text{PFT}$
    \item The round count reaches a multiple of MSL: $(t + 1) \bmod \text{MSL} = 0$
\end{enumerate}

This is formalized using a Boolean switching function $h(t+1)$ that determines whether a new client selection should be made at round $t+1$:

\begin{equation}
h(t+1) =
\begin{cases}
\text{True}, & \text{if } p(t+1) \geq \text{PFT} \text{ or } (t+1) \% \text{MSL} = 0 \\
\text{False}, & \text{otherwise}
\end{cases}
\end{equation}

This mechanism allows FedFiTS to \textbf{dynamically adapt the frequency of client team updates}. Teams that show consistent performance improvements remain active longer, thereby reducing unnecessary communication overhead. Conversely, if performance declines, quicker reselection is enforced to ensure the learning process remains robust and convergent.

To optimize performance, the algorithm prefers \textbf{larger MSL values} (encouraging team stability) and \textbf{smaller PFT values} (allowing quicker reaction to declining performance).

\vspace{2pt}
\noindent{\bf Problem Formulation}

FedFiTS aims to maximize $\theta(.)$, the Quality of Learning \textit{QoL} for a round $t$ or a slot $s$ using fittest clients.
This maximization thus refers to the selection of clients capable of boosting the model to better overall performance. It intrinsically involves the elimination of clients whose impact to the global model is weak or bad, malicious clients, non-compliant with the task or attacked clients,  to avoid propagation of bad parameters to the global model.

The problem is then formulated as follows.

\vspace{-0.2in}
\begin{gather}
	maximize \sum_{t}^{}\sum_{k}^{} \theta_{k}(t)X(k,t) \\
    \text{subject to $score_{k}(t)\geq threshold(t)$ }\\
X(k,t) = 
\left\{
    \begin{array}{lr}
        1, & \text{if $k$ is selected} \\
        0, & \text{other wise }
    \end{array}
\right. 
\end{gather}
where $k$ represents a client, $t$ a round, and $\theta_k$ the client $k$ proximity to the global model (its quality of learning).

As expressed above, FedFiTS builds around its key performance metrics to 1) maximize a reward function expressed as a weighted sum of the clients training  performance $\sum_{t}^{}\sum_{k}^{} \theta_{k}(t)X(k,t)$ and the quality of datasets held by these clients and 2) minimize the number of clients participating in the training process $score_{k}(t)\geq threshold(t)$ with the expectation to increasing the algorithm convergence, increase accuracy and reduce training time. The FedFiTS algorithm is presented below along with the proof of its convergence and performance results revealing its relative performance compared to other key algorithms on different datasets, both images and numerical.  

\vspace{-0.05in}

\subsection{FedFiTS Algorithmic Solution}

The main algorithm of FedFiTS is depicted in Algorithm~\ref{alg:main_algo} while its client updates process is described in Algorithm~\ref{alg:client_update}. At initialization, the algorithm considers $T$ as the total number of rounds and starts with an empty set of selected clients. Free-for-all training allows all available clients in $N_t$ to train the global model at least twice. At round $t=1$, all clients train and send their parameters for aggregation. At round $t=2$, all clients will train but a threshold-aware selection is made to determine the clients which should participate in the aggregation and training for round $t=3$. The team changing process is decided server-side by calculating $h(t+1)$ at round $t$, and that change is performed when $h(t+1)=True$. The selection process follows three steps: (i) calculate the score of each client following Eq.~(\ref{eq:score_equation}), (ii) calculate the threshold using Eq.~(\ref{eq:threshold_equation}), and (iii) select all clients whose score is greater than or equal to the threshold and add them to the current set of  selected clients $S_t$ previously referred to as the ``training and aggregation team''. The aggregation of parameters (weights) is performed in each slot with the clients in $S_t$.

\RestyleAlgo{ruled}
\SetKwComment{Comment}{/* }{ */}
\RestyleAlgo{ruled}
\SetKwComment{Comment}{/* }{ */}

\begin{algorithm}
\caption{FedFiST Algorithm}\label{alg:main_algo}
\SetKwProg{ServerSideProcedure}{FLSAg main Procedure} 
\ServerSideProcedure
    -initialize $w_0$ \\
   initialize $T$ \  \Comment*[r]{Total number of rounds}
   initialize $MSL$ \  \Comment*[r]{Max Slot length}
   initialize $PFT$ \  \Comment*[r]{frequency threshold}
   $p(1) \gets 0$ \  \Comment*[r]{slot monitoring}
   $S_{0} \gets\{\}$ \ \Comment*[r]{set of selected clients}
\ForEach{communication round  $t$=1,2,...$T$}{%
   $N_{t} \gets  (Availaible~clients~at~round~$t$)$\ \Comment*[r]{set of all clients}
   	  
    \If{$h(t)=True$}{ 
    		\ForEach{client $c_{k} \in $N$_{t}$}{%
    	   $w_{k}(t),\theta_k \gets ClientUpdate(c_k,w(t-1))$\ 
        	  } 
        	  \If{$t= 1$}{
            $S_t \gets N_t $\ 
            }
           \Else{
            $S_t\gets \{\}$\ \Comment*[r]{empty St}
        	  \ForEach{client $c_{k} \in $N$_{t}$}{ 
    	        $score_{k}(t) \gets \frac{n_k}{n}\times \alpha +(1-\alpha)\theta_k$\ 
          }
          $threshold_t \gets \frac{\Sigma_{k=1}^{|N_t|}{score_k}}{k} \times (1-\beta)$ \; 
          
          \ForEach{client $c_{k} \in $N$_{t}$}{%
    	   \If{$threshold_t <=score_k $}{ 
            $S_t=S_t \cup{\{c_k\}}$\ \Comment*[r]{Selection}
          }
        } 
           }
      }
    \Else{
    	  $S_t\gets S_{t-1}$ \;
      \ForEach{client $c_{k} \in $S$_{t}$}{%
    	   $w_{k}(t),\theta_k \gets ClientUpdate(c_k,w(t-1))$\
        	  } 	     
    }
    $w(t) \gets \Sigma_{c_k \in S_t}\frac{n_k}{|S_t|}{w_k(t)}$\\
    $\theta(t)\gets \Sigma_{k=1}^{|S_t|}{\theta_k}$\\
    \If{$t>2$ and $\theta(t)<\theta(t-1)$}{  
            $p(t+1)=p(t)+1$
      }
      \Else{
    	$p(t+1)=0$
     }  
    \If{$p(t+1) >= PFT$ or $t+1\%MSL=0$ or $t=1$}{  
            $h(t+1)=True$
      }
      \Else{
    	$h(t+1)=False$
     } 
    }

\end{algorithm}
\vspace{-0.1in}

\begin{algorithm}
\caption{UpdateClient}\label{alg:client_update}
\SetKwFunction{UpdateClient}{ClientUpdate}
\UpdateClient{($c_k,w(t-1)$)}{\\
    $w_{k}(t-1)\gets w(t-1)$\;
    \ForEach{local epoch from 1 to $E$}{
    		\ForEach{batch $b \in B$}{%
	       $w_{k}(t)\gets w(t-1)-\eta_t \nabla(F_k(w(t-1),b)$ \;
    	      }
    	  }
     \If{$t =1 $}{  
            $\theta_k \gets 0$ \Comment*[r]{at round t=1 no angle calculation}
      }
      \Else{
    	$GL_k,GA_k \gets evaluate(w(t-1),testset_k)$ \;
    	$LL_k,LA_k \gets evaluate(w_k(t),testset_k)$ \;
        $\theta_k \gets arcos(\frac{GL_k+LL_k}{\sqrt{(GL_k+GA_k)^2+(LL_k+LA_k)^2}})$\;
     }  
    \KwRet $w_{k},\theta_k$\;
}
\end{algorithm}

\vspace{2pt}
\subsection{FedFiTS in the Selective FL Landscape}

FedFiST is a Federated Learning approach which differs from other approaches on such aspects as the fitness model, client scoring and selection model, and local training process. 

\begin{figure}[htb]
    \centering
    \vspace{-0.1in}
    \includegraphics[width=8.5cm]{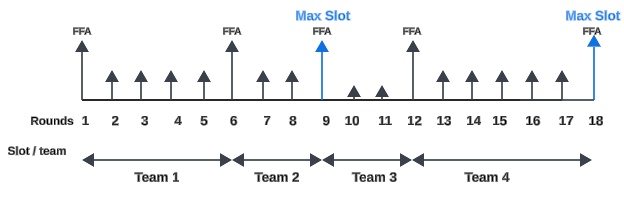}
    \vspace{-0.1in}
    \caption{FedFiST: The Slotted training illustration } 
    \label{fig:fedfast}
    \vspace{-0.05in}
\end{figure}

\begin{figure}[htb]
    \centering
    \vspace{-0.25in}
    \includegraphics[width=9cm]{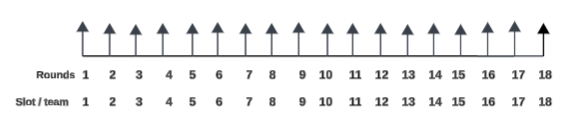}
    \vspace{-0.2in}
    \caption{FedRand and FedPow: Round-based training } 
    \label{fig:others}
  \vspace{-0.05in}
\end{figure}

Figures~\ref{fig:fedfast} and~\ref{fig:others} reveal a key difference between FedFiST and FedRand~\cite{Fedavg} and FedPow~\cite{clientSelectStatPow} 
in terms of clients participation profile in the training and aggregation. For instance, FedFiST uses a {\it slot-based profile} where the clients are evaluated after each slot to form a team of fittest clients. On the other hand, FedRand and FedPow algorithms rely on a {\it round-based profile} where the clients are evaluated at each round to make the fittest client team. Moreover, in FedRand and FedPow, the slot size is of 1  round, whereas in FedFiST, the slots are made of more than one rounds of the FL model execution.

\vspace{-0.2in}

\section{Convergence Analysis of \textsc{FedFiTS}}
\label{sec:conv}

\subsection{Problem Setting and Notation}
Let \(F(w)\!\triangleq\!\sum_{i=1}^N p_i\,f_i(w)\) be the global objective with \(p_i\!\ge\!0\), \(\sum_i p_i=1\).
At round \(t\), a team \(S_t\subseteq\{1,\dots,N\}\) participates (FedFiTS uses slotted teams).
Each \(i\in S_t\) runs \(E\) local SGD steps with step size \(\eta_\ell\) on minibatches \(\xi\):
\begin{equation}
w_{i}^{(e+1)} \;=\; w_{i}^{(e)} - \eta_\ell\, g_i\!\big(w_{i}^{(e)},\xi\big),\qquad
e=0,\dots,E-1,\ \ w_{i}^{(0)}=w_t .
\end{equation}
The server forms a (possibly robust) aggregate gradient surrogate \(\bar g_t\) from \(\{w_i^{(E)}\}_{i\in S_t}\) and updates
\begin{equation}
w_{t+1} \;=\; w_t \;-\; \eta_g\,\bar g_t ,
\end{equation}
with global step size \(\eta_g>0\). Write \(\tilde g_{i,t}\) for the effective client gradient after \(E\) local steps, and define
\begin{equation}
\bar g_t \;=\; \mathcal{A}\big(\{\alpha_{i,t}\,\tilde g_{i,t}\}_{i\in S_t}\big), \quad
\alpha_{i,t}\!\ge\!0,\ \sum_{i\in S_t}\alpha_{i,t}=1,
\end{equation}
where \(\mathcal{A}(\cdot)\) is a robust aggregator (FedAvg/median/trimmed mean/Krum, etc.).

\subsection{Assumptions}
\begin{itemize}
\item[(A1)] \textbf{Smoothness:} Each \(f_i\) is \(L\)-smooth:
\(\|\nabla f_i(x)-\nabla f_i(y)\|\le L\|x-y\|\).
\item[(A2)] \textbf{Stochastic noise:} \(\mathbb{E}[g_i(w,\xi)\mid w]=\nabla f_i(w)\) and
\(\mathbb{E}\|g_i(w,\xi)-\nabla f_i(w)\|^2\le \sigma^2\).
\item[(A3)] \textbf{Client dissimilarity:}
\(\sum_{i=1}^N p_i \|\nabla f_i(w)-\nabla F(w)\|^2\le \zeta^2\), \(\forall w\).
\item[(A4)] \textbf{Selection floors \& bias:} FedFiTS’ quotas/exploration ensure
\(\Pr(i\!\in\!S_t)\ge p_{\min}>0\) and the selection bias
\begin{equation}
\varepsilon_{\mathrm{sel},t}\;\triangleq\;\Big\|\sum_{i\in S_t}\alpha_{i,t}\nabla f_i(w_t)-\sum_{i=1}^N p_i\nabla f_i(w_t)\Big\|
\end{equation}
satisfies \(\mathbb{E}\,\varepsilon_{\mathrm{sel},t}^2\le \varepsilon_{\mathrm{sel}}^2\).
\item[(A5)] \textbf{Robust aggregation error:}
\(\mathbb{E}\|\bar g_t - \sum_{i\in S_t}\alpha_{i,t}\tilde g_{i,t}\|^2\le \varepsilon_{\mathrm{agg}}^2\).
\item[(A6)] \textbf{Local/slot drift:}
\(\mathbb{E}\|\tilde g_{i,t}-\nabla f_i(w_t)\|^2\le \varepsilon_{\mathrm{loc}}^2\), and slot staleness (fixed team within a slot) yields
\(\mathbb{E}\,\varepsilon_{\mathrm{slot},t}^2\le \varepsilon_{\mathrm{slot}}^2\).
\item[(A7)] \textbf{Stepsizes:} \(\eta_g\le 1/(2L)\); choose \(\eta_\ell\) such that
\(\varepsilon_{\mathrm{loc}}^2=O\!\big(E\,\eta_\ell^2(\sigma^2+\zeta^2)\big)\).
\end{itemize}
Collect residuals in
\begin{equation}
\mathcal{R}\;\triangleq\;\varepsilon_{\mathrm{sel}}^2+\varepsilon_{\mathrm{agg}}^2+\varepsilon_{\mathrm{loc}}^2+\varepsilon_{\mathrm{slot}}^2 .
\end{equation}

\subsection{One-Step Descent}
\noindent\textbf{Lemma 1 (Expected Descent).}
\emph{Under (A1)–(A7), there exist constants \(c_1,c_2,c_3=O(1)\) such that}

\begin{multline}
\mathbb{E}\big[F(w_{t+1})\big] \;\le\;
\mathbb{E}\big[F(w_t)\big]
-\frac{\eta_g}{2}\,\mathbb{E}\|\nabla F(w_t)\|^2 \\
+ \eta_g\,\Big( c_1\,\sigma^2 + c_2\,\zeta^2 + c_3\,\mathcal{R} \Big)
\end{multline}

\emph{Sketch.} Apply \(L\)-smoothness to \(F\) at \(w_{t+1}=w_t-\eta_g\bar g_t\), decompose \(\bar g_t\) into the ideal global gradient plus (zero-mean) stochastic noise, heterogeneity, selection/aggregation errors, and local/slot drift; then use Young’s inequality and \(\eta_g\le 1/(2L)\).

\subsection{Non-Convex Stationarity}
\noindent\textbf{Theorem 1 (Stationarity Bound).}
\emph{Let \(F^\star\!\triangleq\!\inf_w F(w)\).
Under (A1)–(A7) with \(\eta_g\le 1/(2L)\),}

\begin{multline}
\min_{0 \le t < T} \ \mathbb{E}\|\nabla F(w_t)\|^2 \;\le\;
\frac{2\big(F(w_0) - F^\star\big)}{\eta_g T} \\
+ 2\Big( c_1\,\sigma^2 + c_2\,\zeta^2 + c_3\,\mathcal{R} \Big)
\end{multline}

\emph{Proof.} Sum Lemma~1 over \(t=0,\dots,T-1\) and rearrange.

\subsection{Linear Rate under Polyak--\L{}ojasiewicz (PL)}
\noindent\textbf{Assumption (PL).}
There exists \(\mu>0\) such that \(\tfrac{1}{2}\|\nabla F(w)\|^2 \ge \mu\,(F(w)-F^\star)\) for all \(w\).

\noindent\textbf{Corollary 1 (Linear Convergence to a Neighborhood).}
\emph{Under (A1)–(A7) and PL, with \(\eta_g\le 1/(2L)\),}

\begin{multline}
\mathbb{E}\big[F(w_{t+1}) - F^\star\big] \;\le\;
(1 - \mu \eta_g)\,\mathbb{E}\big[F(w_t) - F^\star\big] \\
+ \eta_g\,\tilde{c}\,\big(\sigma^2 + \zeta^2 + \mathcal{R}\big)
\end{multline}

\emph{for some \(\tilde c=O(1)\). Hence, for all \(t\),}
\begin{equation}
\mathbb{E}\big[F(w_{t})-F^\star\big]
\;\le\;
(1-\mu\eta_g)^t\,\big(F(w_0)-F^\star\big)
+\frac{\tilde c}{\mu}\,\big(\sigma^2+\zeta^2+\mathcal{R}\big).
\end{equation}

\subsection{Remarks}
\begin{itemize}
\item \textbf{Selection floors / explore--exploit} in FedFiTS bound \(\varepsilon_{\mathrm{sel}}^2\) and improve gradient representativeness.
\item \textbf{Multi-objective fitness} favors larger/consistent clients, effectively reducing \(\zeta^2\) and noise.
\item \textbf{Robust aggregation} tightens \(\varepsilon_{\mathrm{agg}}^2\) under outliers/poisoning.
\item \textbf{Adaptive slots (MSL/PFT)} cap \(\varepsilon_{\mathrm{slot}}^2\) by early reselection when plateaus are detected.
\end{itemize}

\section{Dynamic $\alpha$ parameter determination}\label{sec:dynamic}
\subsection{Dynamic $\alpha$ determination principle}
The \textbf{FedFiST} scoring function has been formulated and expressed by Eq.~(\ref{eq:score_equation}). To find the optimal $\alpha$, we express each client's score function in terms of $\alpha_{k}$ as $score(\alpha_{k},t)=(q_{k}-\theta_{k})\times \alpha_{k} +\theta_{k} $, which results in a line segment with $\alpha_{k}$ bounded between $[0, 1]$. Finding the $\alpha_{k}$ that maximizes the score for the client is equivalent to finding the maximum of the segment, implying the calculation of its values at the boundaries. In our case, 
\begin{gather}
\alpha_{k} = 
\left\{
    \begin{array}{lr}
        1, & \text{if } q_{k} > \theta_{k}\\
        0, & \text{if } q_{k} \leq \theta_{k}
    \end{array}
\right.
\end{gather}
After determining all the $\alpha_{k}$ values, which are the abscissas maximizing the clients’ score function, we need to find the unique $\alpha$ value to be used for client selection. 
The principle of function maximization would suggest using the $\alpha_{k}$ that maximizes the segment with the highest maximum value as the abscissa of the global function. 
However, in our context, it is very risky to choose 1 or 0 as values due to (1) exposure to attacks when $\alpha$ = 1, and (2) strict exclusion of all clients that are not performing well no matter what sizes they have when $\alpha = 0$. 
To avoid these extremes values, we use the average of the $\alpha_{k}$, as it theoretically provides information on the proximity to the maximum.
\begin{equation}
    \alpha= \sum_{k}^{} \alpha_{k}
\end{equation}
This dynamic $\alpha$ is subject to the quality and the size of the clients. Whenever $\sum_{k=1}^{K}1(q_{k}>\theta_{k}) > \sum_{k=1}^{K}1(q_{k}<\theta_{k})$ $\alpha>0.5$ and $\alpha<0.5$ otherwise.

```latex
\section{Experimental Evaluation}

This section evaluates the performance of \textit{FedFiTS} in terms of accuracy, convergence speed, robustness to poisoning, communication efficiency, and fairness considerations. The experiments are conducted across multiple datasets and compared against standard and selective FL baselines.

\subsection{Experimental Setup}
We evaluate \textbf{FedFiTS} on three categories of datasets: 
(i) \textbf{Medical imaging datasets}, including the X-ray Pneumonia Detection dataset; 
(ii) \textbf{Vision benchmarks}, namely MNIST and Fashion-MNIST; and 
(iii) \textbf{Tabular datasets}, including the Crop Recommendation dataset which demonstrates cross-domain applicability. 
Data distributions are partitioned under non-IID settings to reflect real-world heterogeneity. 

Baselines include \textbf{FedAvg}, \textbf{FedRand}, and \textbf{FedPow}, representing standard and selective FL approaches. In all cases, we simulate client unreliability and adversarial behavior through data and model poisoning attacks. Hyperparameters are tuned to ensure fairness across methods.

\subsection{Performance Metrics}
Evaluation is based on the following dimensions:
\begin{itemize}
    \item \textbf{Accuracy and Convergence Speed}: classification accuracy over training rounds and rate of convergence.  
    \item \textbf{Robustness to Attacks}: degradation under poisoning and adversarial scenarios.  
    \item \textbf{Communication Efficiency}: per-round uplink/downlink cost relative to FedAvg and FedPow.  
    \item \textbf{Fairness Considerations}: analysis of participation ratios and client inclusion balance, serving as a proxy for fairness. While explicit fairness metrics (e.g., group accuracy balance, disparity reduction) are not directly measured here, \textbf{FedFiTS}'s selective strategy inherently promotes fairness by dynamically including reliable but minority clients. We highlight this as a future direction.  
\end{itemize}

\subsection{Results on Imaging and Vision Datasets}
\begin{figure}[ht]
    \centering
    \includegraphics[scale=0.50]{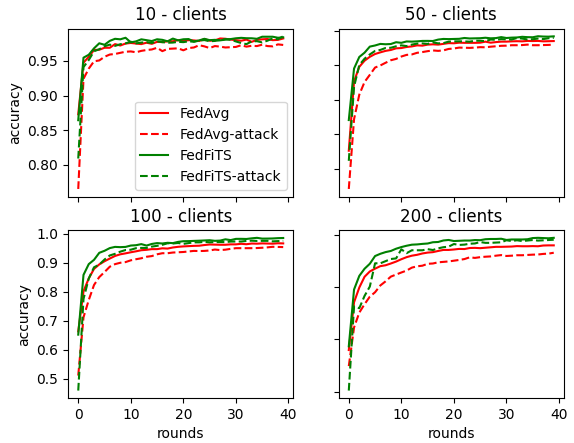}
    \caption{Accuracy comparison of FedAvg and \textbf{FedFiTS} under normal and attack modes on MNIST.}
    \label{fig:comparison_FedAVG_FedFast}
\end{figure}

\begin{figure}[h]
    \centering
    \includegraphics[scale=0.55]{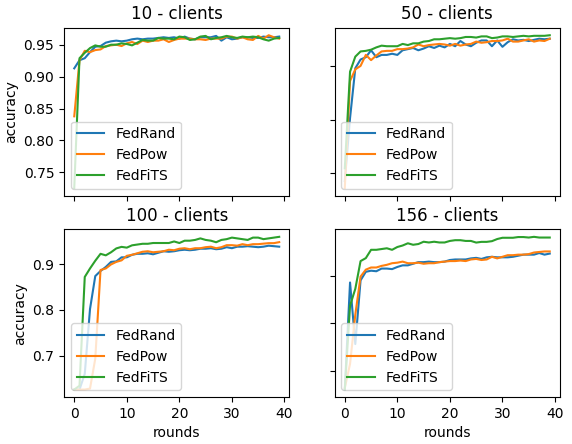}
    \caption{X-ray dataset performance: normal mode.}
    \label{fig:xrays_normal}
\end{figure}

\begin{figure}[h]
    \centering
    \includegraphics[scale=0.55]{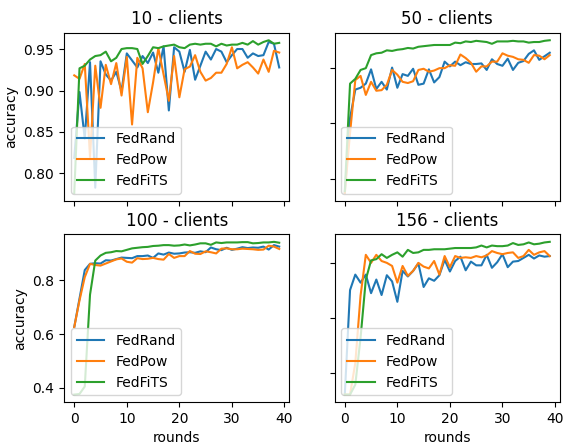}
    \caption{X-ray dataset performance: attack mode.}
    \label{fig:xrays_normal_attack}
\end{figure}

Figures~\ref{fig:comparison_FedAVG_FedFast}--\ref{fig:auto_mm}, together with the corresponding tables, present the comparative evaluation of \textit{FedFiTS} against baseline algorithms including FedAvg, FedRand, and FedPow. The results span multiple datasets—MNIST, Fashion-MNIST, and X-ray Pneumonia—covering both vision benchmarks and real-world medical imaging tasks. 

\subsection*{Results on MNIST Dataset}
We first compare \textbf{FedFiTS} (slot size = 1) with FedAvg ($c=1.0$). Both use all available clients per round.  
Training/testing: 10,000 / 2,000 MNIST images. Teams of 10, 50, 100, and 200 clients, partitioned using Dirichlet distributions with $\alpha = 0.3, 0.2, 2.0, 1.0$.  

\begin{table}[ht]
\centering
\caption{\textbf{FedFiTS} (slot size = 1) vs. FedAvg ($c=1.0$) on MNIST: Accuracy and Execution Time}
\label{table:result_table}
\begin{tabular}{ |c|c|c|c|c|c| } 
\hline
Data  & $K$   & \makecell{FedAvg\\Acc} & \makecell{\textbf{FedFiTS}\\Acc} & \makecell{FedAvg\\Time (s)} & \makecell{\textbf{FedFiTS}\\Time (s)} \\
\hline
Normal & 10  & 0.982 & \textbf{0.984} & 966.0 & \textbf{959.3} \\ 
       & 50  & 0.971 & \textbf{0.983} & 1881.1 & \textbf{1695.8} \\ 
       & 100 & 0.967 & \textbf{0.986} & 3304.5 & \textbf{2745.0}\\ 
       & 200 & 0.959 & \textbf{0.989} & 6257.6 & \textbf{5133.5}\\ 
\hline
Attack & 10  & 0.973 & \textbf{0.979} & 978.6 & \textbf{905.2} \\ 
       & 50  & 0.961 & \textbf{0.979} & 1987.3 & \textbf{1769.0} \\ 
       & 100 & 0.958 & \textbf{0.980} & 3892.5 & \textbf{2737.6}\\ 
       & 200 & 0.932 & \textbf{0.981} & 6021.2 & \textbf{5083.3}\\ 
\hline
\end{tabular}
\end{table}

\noindent\textbf{Findings:}  
\textbf{FedFiTS} consistently outperforms FedAvg in both normal and attack modes, achieving higher accuracy and lower execution time. The performance gap widens as the number of clients increases, and resilience against poisoning attacks is clearly demonstrated.

\subsection*{Results on X-Ray Dataset}
We evaluated \textbf{FedFiTS}, FedRand, and FedPow on the Pneumonia X-ray dataset with 3,792 training and 943 test samples. Patients were grouped into clients to reflect realistic hospital-based distributions.  

\begin{table}[ht]
\centering
\caption{Patient Dispatch and $m$, $d$ Values for X-ray Experiments}
\begin{tabular}{|c|c|c|c|c|}
\hline
K & \# Patients & \# Clients & $m$ & $d$ \\
\hline
10 & 280 & 10 & 7 & 9 \\
50 & 56  & 50 & 32 & 41 \\
100& 28  & 100 & 58 & 82 \\
156& 18  & 156 & 113 & 124 \\
\hline
\end{tabular}
\label{table:patient_dispatch}
\end{table}

\begin{center}
\begin{scriptsize}
\begin{table}[ht]
\centering
\caption{FedRand vs FedPow vs \textbf{FedFiTS} on X-rays: Accuracy and Execution Time}
\label{table:result_table_xrays}
\vspace{-0.05in}
\begin{tabular}{ |c|c|c|c|c| } 
\hline
 & &\multicolumn{3}{|c|}{Accuracy} \\
 \hline
                    Case  & K  &  FedRand & FedPow & \textbf{FedFiTS}  \\
\hline
\multirow{1}{2.5em}{Normal} & 10  & 0.960  & 0.950  & \textbf{0.964} \\ 
                            & 50   & 0.950 & 0.950  & \textbf{0.955}   \\ 
                            & 100 & 0.937 &  0.947 & \textbf{0.960} \\ 
                            & 156 & 0.946 & 0.950 & \textbf{0.968} \\ 
\hline
\hline
\multirow{1}{2.5em}{Attack} & 10  & 0.927  & 0.930  & \textbf{0.957} \\ 
                            & 50  & 0.927  & 0.923  & \textbf{0.950}  \\ 
                            & 100 & 0.925  &  0.916 & \textbf{0.940} \\ 
                            & 156 & 0.913  & 0.913  & \textbf{0.940}  \\ 
\hline
\end{tabular}

\vspace{0.35em}

\begin{tabular}{ |c|c|c|c|c| } 
\hline
 & &\multicolumn{3}{|c|}{Execution time/s}\\
 \hline
                    Case  & K   & FedRand & FedPow & \textbf{FedFiTS} \\
\hline
\multirow{1}{2.5em}{Normal} & 10   & 958  & 1179  & 1153 \\ 
                            & 50   & 1813 & 1895  & 1889  \\ 
                            & 100  & 2315 & 2599  & 2420\\ 
                            & 156  & 4009 & 4900  & 4018 \\ 
\hline
\hline
\multirow{1}{2.5em}{Attack} & 10   & 918  & 1101  & 1058 \\ 
                            & 50   & 1705 & 1797  & 1790  \\ 
                            & 100  & 2941 & 3085  & 2931\\ 
                            & 156  & 3451 & 3719  & 3440 \\ 
\hline
\end{tabular}
\end{table}
\end{scriptsize}
\end{center}

\noindent\textbf{Findings:} \textbf{FedFiTS} consistently surpasses FedRand and FedPow in accuracy under both normal and attack conditions. Execution times are comparable, with \textbf{FedFiTS} gaining efficiency as team size increases.

\subsubsection*{Summary of Results on Vision and Imaging Datasets}
Overall, the experimental results on MNIST, Fashion-MNIST, and X-ray datasets confirm the effectiveness of \textit{FedFiTS} in vision and medical imaging tasks. Under non-IID data partitions, \textbf{FedFiTS} consistently achieves higher accuracy and faster convergence compared to FedAvg, FedRand, and FedPow. In adversarial settings with data poisoning, its selective aggregation mechanism effectively filters out compromised clients, leading to greater robustness and stability across rounds. Taken together, these results demonstrate that \textbf{FedFiTS} provides significant improvements in accuracy, convergence speed, and resilience relative to existing FL approaches when applied to image-based learning tasks.

\subsection{Results on Tabular Crop Dataset}

To further assess the cross-domain adaptability of \textit{FedFiTS}, we evaluate its performance on the Crop Recommendation dataset~\cite{cropdataset}, a structured tabular dataset that differs substantially from the vision-based tasks considered earlier. The dataset consists of 22,000 samples with 22 features per record, including key agricultural indicators such as nitrogen, phosphorus, potassium, soil pH, temperature, rainfall, and humidity. Each record is labeled with one of 22 crop types, making the task a multiclass classification problem. This dataset provides a challenging evaluation scenario as it involves heterogeneous feature types and imbalanced class distributions, reflecting conditions often encountered in Internet of Things (IoT) and precision agriculture applications.

Data were partitioned into multiple clients following a non-IID distribution to simulate realistic deployment across farms or agricultural regions with varying soil and climate conditions. Similar to previous experiments, we compared \textit{FedFiTS} against FedAvg, FedRand, and FedPow under increasing numbers of clients.

\begin{figure}[h]
    \centering
    \includegraphics[scale=0.55]{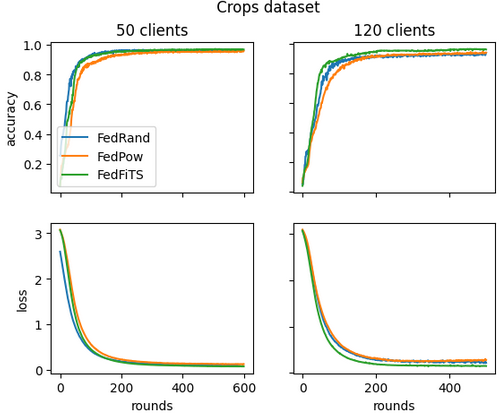}
    \caption{Comparison on Crop dataset: \textit{FedFiTS} consistently outperforms baseline algorithms, with performance gains widening as the number of clients increases.}
    \label{fig:cropdataset}
\end{figure}

Figure~\ref{fig:cropdataset} illustrates the results. \textit{FedFiTS} not only maintains its accuracy advantage but also demonstrates faster convergence relative to FedAvg and other selective FL approaches. Importantly, the performance gap widens as the number of clients increases, indicating that \textit{FedFiTS} scales more effectively under high client participation scenarios. This is attributed to its dynamic client scoring mechanism, which ensures that clients with reliable data and consistent performance are prioritized, thereby mitigating the negative impact of unreliable or biased clients.

These results confirm that \textit{FedFiTS} is not restricted to image-based healthcare tasks but generalizes well to structured tabular learning. Such adaptability underscores its potential applicability in a wide range of domains, including agriculture, IoT-based environmental monitoring, and smart city analytics. The findings highlight the robustness and flexibility of \textit{FedFiTS} in handling diverse data modalities while preserving its core advantages of improved accuracy, faster convergence, and enhanced robustness.
\vspace{0.05in}

\subsection{Impact of Hyperparameter Tuning on Performance}  
\begin{figure}[h]
    \centering
    \includegraphics[scale=0.55]{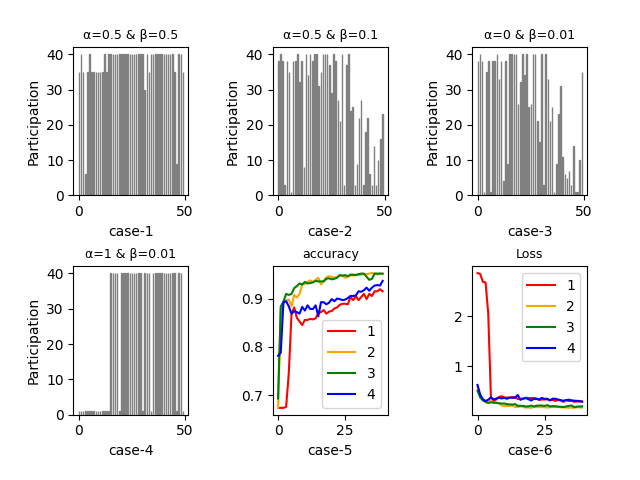}
    \caption{Impact of tuning $\alpha$ and $\beta$ on convergence under compromised client scenarios.}
    \label{fig:xrays_hyper_parameters}
\end{figure}

The convergence behavior of \textit{FedFiTS} is strongly influenced by the choice of the hyperparameters $\alpha$ and $\beta$.  
As defined in Eqs.~(\ref{eq:score_equation}) and (\ref{eq:threshold_equation}), $\alpha$ controls the trade-off between client performance and data quality, while  $\beta$ regulates the openness of the threshold used to differentiate fittest clients from borderline ones.  
Figure~\ref{fig:xrays_hyper_parameters} illustrates the effects of varying $\alpha$ at boundary values  
($\alpha = 0, 0.5, 1$) combined with different $\beta$ values under conditions where some clients contribute compromised data.  
We examine four representative cases:  
\begin{itemize}
    \item \textit{Case 1} ($\alpha=0.5$, $\beta=0.5$): balances performance and data quality with 50\% openness.  
    \item \textit{Case 2} ($\alpha=0.5$, $\beta=0.1$): maintains the same balance as Case 1 but with reduced openness (10\%).  
    \item \textit{Case 3} ($\alpha=0$, $\beta=0.01$): ignores data quality, selecting only high-performing clients.  
    \item \textit{Case 4} ($\alpha=1$, $\beta=0.01$): ignores performance, prioritizing clients with large datasets.  
\end{itemize}

The results indicate clear trade-offs:  
\begin{itemize}
    \item \textbf{Case 2} achieves the best convergence by limiting the participation of compromised clients, thus ensuring balanced representation.  
    \item \textbf{Case 1}, with higher openness, allows more compromised clients into the training team, degrading performance.  
    \item \textbf{Case 4} overemphasizes large datasets, repeatedly selecting the same clients and excluding smaller ones.  
    \item \textbf{Case 3} favors only high-performing clients, disregarding smaller but potentially valuable datasets.  
\end{itemize}

Overall, the accuracy (Case 5) and loss (Case 6) curves confirm that Case 2 yields superior convergence by proportionally balancing client participation according to the actual data distribution.  

Figure~\ref{fig:xrays_hyper_parameters} presents the results of testing the sensitivity of $\alpha$ (data-performance trade-off) and $\beta$ (threshold openness). 

\subsection{Hyperparameter Sensitivity Analysis}  

The hyperparameters $\alpha$ and $\beta$ play a central role in balancing client participation, robustness, and convergence in \textit{FedFiTS}. As defined in the scoring and threshold equations, $\alpha$ regulates the trade-off between data quality and model performance, while $\beta$ controls the openness of the client selection threshold.  

\begin{figure}[h]
    \centering
    \includegraphics[scale=0.50]{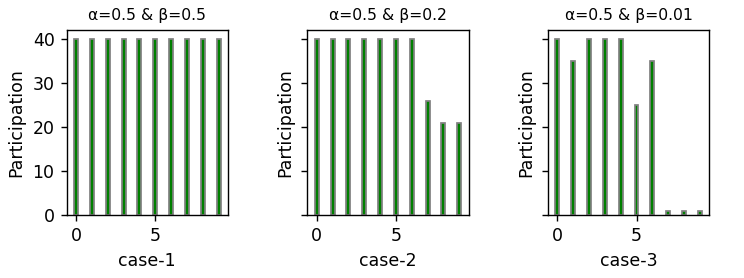}
    \caption{Tuning $\beta$: \textit{FedFiTS} prevents compromised clients—specifically the last four— from joining the training team.}
    \label{fig:tuning_beta}
\end{figure}

Figure~\ref{fig:tuning_beta} illustrates the effect of tuning $\beta$. Reducing $\beta$ (e.g., from 0.5 to 0.01) effectively excludes compromised clients from the training team, thereby improving robustness. Higher $\beta$ values allow greater openness in participation but risk admitting poisoned updates, while lower values favor stability by restricting selection to reliable clients. Balancing $\alpha$ ensures that both strong performance and sufficient data quality contribute to aggregation, preventing bias toward one factor.  

\begin{figure}[ht]
    \centering
    \vspace{-0.2in}
    \includegraphics[scale=0.55]{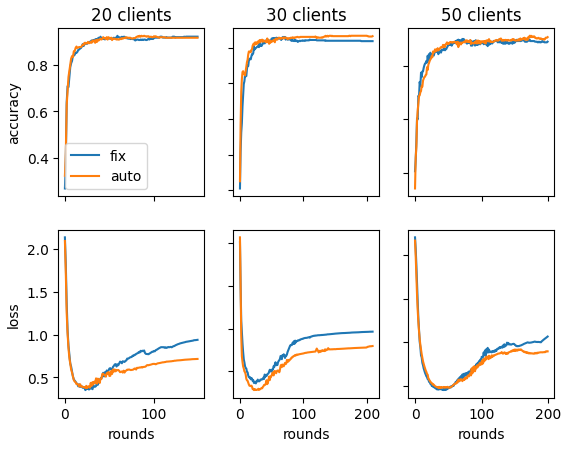}
    \vspace{-0.03in}
    \caption{Comparison of fixed $\alpha=0.5$ and dynamically determined $\alpha$ on MNIST dataset.}
    \label{fig:auto_mt}
    \vspace{+0.1in}
\end{figure}

\begin{figure}[ht]
    \centering
    \vspace{-0.2in}
    \includegraphics[scale=0.55]{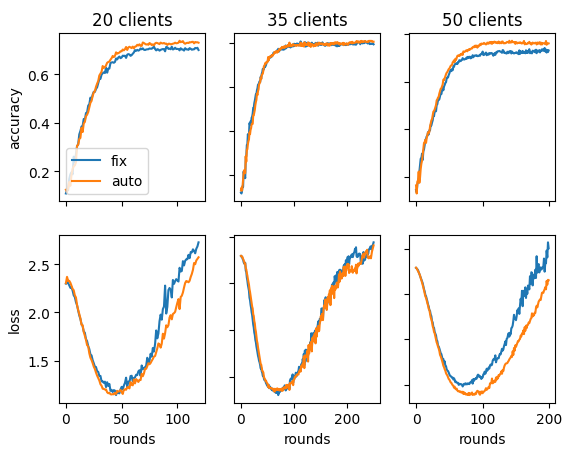}
    \vspace{-0.03in}
    \caption{Comparison of fixed $\alpha=0.5$ and dynamically determined $\alpha$ on MNIST-M dataset.}
    \label{fig:auto_mm}
    \vspace{-0.1in}
\end{figure}

To further assess hyperparameter sensitivity, we compared \textit{FedFiTS} under two strategies for $\alpha$: (i) a fixed setting with $\alpha=0.5$, and (ii) a dynamic configuration where $\alpha$ is automatically recalculated at each round or slot. Experiments across multiple client-team configurations on MNIST and MNIST-M confirm that the dynamic approach generally outperforms the fixed configuration (Figures~\ref{fig:auto_mt} and \ref{fig:auto_mm}), though not in all cases. Its strength lies in the adaptive rebalancing of priorities between data quantity and model proximity across rounds, enabling a more effective optimization trajectory for the global model.  

\noindent\textbf{Proxy fairness evaluation:}  
To quantify fairness, we measured the \emph{participation ratio}, defined as the fraction of unique clients selected at least once during training. Table~\ref{table:fairness_proxy} summarizes the results.  

\begin{table}[h]
\centering
\caption{Client participation ratio (\%) under different settings.}
\label{table:fairness_proxy}
\renewcommand{\arraystretch}{1.1}
\begin{tabular}{|c|c|}
\hline
\textbf{Configuration} & \textbf{Participation (\%)} \\
\hline
FedAvg    & 45 \\
FedPow    & 52 \\
FedFiTS ($\alpha=0.5,\beta=0.5$) & 65 \\
FedFiTS ($\alpha=0.5,\beta=0.1$) & 78 \\
FedFiTS (Dynamic $\alpha$)       & 82 \\
\hline
\end{tabular}
\end{table}

The results indicate that \textit{FedFiTS} achieves significantly higher client inclusion compared to FedAvg and FedPow. In particular, the configuration with $\alpha=0.5, \beta=0.1$ yields balanced participation while filtering out unreliable clients, whereas dynamic $\alpha$ provides the highest ratio by adaptively adjusting selection priorities across rounds. These findings support the interpretation that \textit{FedFiTS} promotes fairness by preventing minority clients from being systematically excluded from training. This aligns with prior fairness-focused studies in FL~\cite{wang2020fair,li2021fair}, which emphasize balancing participation and mitigating disparities among heterogeneous clients.  

\noindent\textbf{Summary:}  
Overall, these experiments confirm that $\alpha$ and $\beta$ decisively influence accuracy, robustness, and fairness in \textit{FedFiTS}. Static settings may overemphasize dataset size or client performance, leading to biased participation and degraded convergence. Moderate tuning achieves balanced participation and better convergence, while dynamic adjustment of $\alpha$ further improves adaptability by recalculating priorities under evolving training conditions. Together, these results demonstrate that careful hyperparameter sensitivity management is crucial for robust and fair deployment of \textit{FedFiTS}.  

\subsection{Trust and Transparency in Federated Learning-Based Healthcare Systems}  

Trust is a fundamental factor in the adoption of AI-driven medical solutions. In federated learning, {lack of direct control over data sources and model updates} can lead to concerns regarding reliability, explainability, and fairness. Clinicians and healthcare institutions require {clear insights into how AI models make decisions}, especially in critical fields such as radiology, pathology, and diagnostics.  To enhance trust in federated AI-driven healthcare, \textit{FedFiTS} can incorporate:  
\begin{itemize}
    \item \textbf{Explainability Mechanisms}: Providing interpretable insights into model outputs (e.g., why a chest X-ray is classified as pneumonia).  
    \item \textbf{Uncertainty Estimation}: Displaying confidence scores alongside diagnoses to help clinicians assess the reliability of AI-generated predictions.  
    \item \textbf{Auditable AI Pipelines}: Implementing \textit{transparency measures} that allow medical experts to validate FL training data sources and model changes over time.  
\end{itemize}

Future research should focus on how {clinicians’ trust in FL models evolves over time} and how {model transparency impacts medical decision-making} in real-world settings.  

\subsection{Fairness and Bias in Federated Learning Models}  

Bias and fairness remain central challenges in federated learning, particularly for healthcare and other sensitive domains where decisions directly affect diverse populations. Due to the non-IID nature of client data, global models may inadvertently prioritize majority clients while under-representing minority participants. This imbalance reduces predictive accuracy for certain groups and risks reinforcing systemic inequities.  \textit{FedFiTS} mitigates these risks by integrating fairness-aware client selection into its aggregation process. Clients are dynamically evaluated based on both data quality and performance stability, ensuring that reliable but minority clients are not excluded from training. This prevents bias amplification and promotes more equitable contributions to the global model.  While we did not benchmark with standard fairness metrics, we considered \emph{client participation ratio} as a proxy. Results (Figures~\ref{fig:tuning_beta}, \ref{fig:xrays_hyper_parameters}) show that unlike FedAvg and FedPow, which often over-select majority clients, \textit{FedFiTS} adaptively filters out unreliable updates while maintaining diverse participation. This balanced inclusion serves as evidence that \textit{FedFiTS} promotes fairness alongside robustness and accuracy.  

By combining fairness-aware selection with proxy fairness evaluation, \textit{FedFiTS} demonstrates potential as a lightweight yet equitable approach to federated learning in real-world deployments.  
The \textit{FedFiTS} algorithm aims to optimize the training performance with a goal to increase accuracy, and reduce losses and execution time of FL model while ensuring that the algorithm converges.  

\section{Conclusion and Future Research Directions}  

This paper introduced \textit{FedFiTS}, a selective federated learning algorithm designed to enhance accuracy, convergence speed, and security in healthcare AI. By employing a fitness-based client selection strategy, \textit{FedFiTS} ensures that only high-quality models and data sources contribute to the global process, thereby improving robustness against data poisoning. Experimental evaluations on X-ray Pneumonia and Crop Recommendation datasets demonstrated that \textit{FedFiTS} achieves higher accuracy, faster convergence, and stronger resilience than FedAVG, FedRand, and FedPow. The algorithm further improves efficiency through slotted training and adaptive client selection, highlighting its potential for clinical deployment and broader healthcare applications.  

Looking ahead, several directions remain open. First, trust and transparency must be strengthened through explainability and communication strategies that enhance clinician confidence. Second, fairness requires explicit evaluation using metrics such as group accuracy balance and disparity reduction, supported by safeguards like demographic-aware audits and diversity-driven client selection. Third, security should extend beyond poisoning resistance to address backdoor attacks and gradient inversion, potentially via differential privacy, secure aggregation, or trusted execution environments. Finally, scalability and cross-domain deployment will demand efficient communication-compression strategies and adaptive scheduling to support large-scale IoT and smart city settings.  

By advancing along these directions, federated learning systems like \textit{FedFiTS} can evolve into scalable, trust-aware, and fairness-oriented solutions suitable for healthcare and beyond.  

\vspace{0.05in}
\noindent{\bf Acknowledgments:}
This work is in part supported by Telkom/ Openserve/Aria Technologies Centre-of-Excellence at Computer Science at University of the Western Cape, the University of Missouri South African Education Program (UMSAEP) program, and the US National Science Foundation (NSF) project CNS-2008878 on "Robust Federated Learning for IoT."

\end{document}